\documentclass[letterpaper, 10 pt, journal, twoside]{packages/IEEEtran}

\IEEEoverridecommandlockouts                              % This command is only needed if 
\usepackage{graphics} % for pdf, bitmapped graphics files
\usepackage{epsfig} % for postscript graphics files
\usepackage{mathptmx} % assumes new font selection scheme installed
\usepackage{times} % assumes new font selection scheme installed
\usepackage{amsmath} % assumes amsmath package installed
\usepackage{amssymb}  % assumes amsmath package installed
\usepackage{bm}

\usepackage{booktabs}
\usepackage{caption}
\usepackage{etoolbox}
\usepackage{graphicx}
\usepackage{multicol}
\usepackage{multirow}
\usepackage{pifont}
\usepackage{xcolor}
\usepackage{hyperref}

\hypersetup{
	colorlinks=true,
	linkcolor=blue,
	filecolor=magenta,      
	urlcolor=magenta,
	citecolor=blue,
}

\newcommand{\PAR}[1]{\vskip4pt \noindent{\bf #1~}}

\begin{document}
	\title{NeRF-VO: Real-Time Sparse Visual Odometry \\ with Neural Radiance Fields}
	
	\author{Jens Naumann$^{1}$, Binbin Xu$^{2}$, Stefan Leutenegger$^{1}$, and Xingxing Zuo$^{1,3,\ast}$% <-this % stops a space
%		\\ \url{https://xingxingzuo.github.io/nerfvo}
 \\ \href{https://xingxingzuo.github.io/nerfvo}{https://xingxingzuo.github.io/nerfvo} 
		\thanks{This work was partially supported by the Munich Center of Machine Learning (MCML).}
		\thanks{$^{1}$Jens Naumann, Stefan Leutenegger, and Xingxing Zuo are with Technical University of Munich, 80333 München, Germany. {\tt\footnotesize \{jens.naumann, stefan.leutenegger, xingxing.zuo\}@tum.de}}%
		\thanks{$^{2}$ Binbin Xu is with University of Toronto,  M3H 5T6 Toronto, Canada. {\tt\footnotesize binbin.xu@utoronto.ca}}%
		\thanks{$^{3}$ Xingxing Zuo is also with California Institute of Technology, Pasadena, CA 91125, USA.}
		\thanks{$^\ast$ {Corresponding author: Xingxing Zuo.}}
%		\thanks{Digital Object Identifier (DOI): see top of this page.}
	}
	
	\newcommand{\insertfig}{
		\begin{center}
			\centering
			\captionsetup{type=figure}
			\setcounter{figure}{0}
			\includegraphics[width=\textwidth]{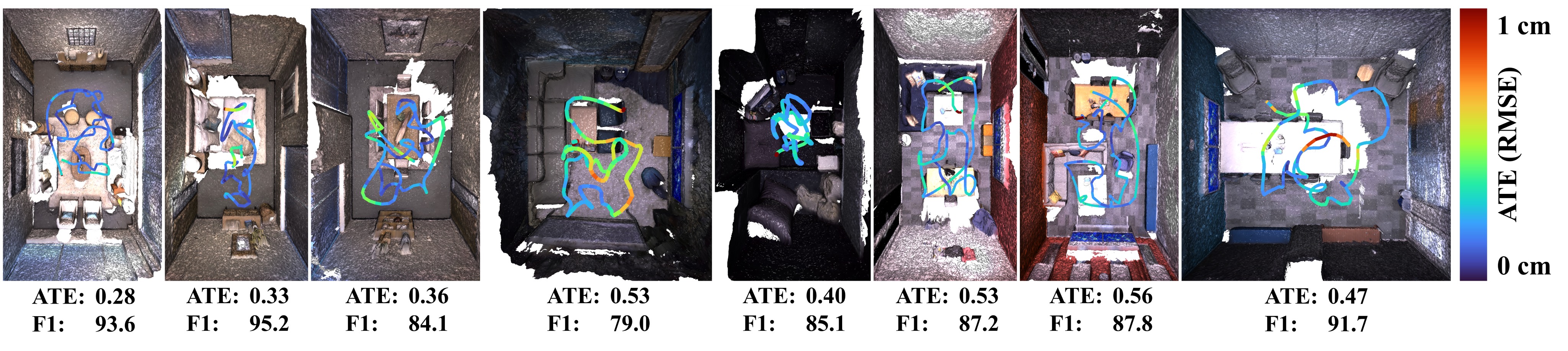}
			\captionof{figure}{\textbf{3D reconstruction and camera tracking results on Replica~\cite{replica19arxiv, Sucar:etal:ICCV2021}.} Meshes rendered from optimized neural radiance fields. Scenes from left to right: room0-2, and office0-4. Quantitative evaluations of pose estimation and 3D reconstruction are reported in cm (ATE RMSE) and \% (F1 score [\textless 5cm]). NeRF-VO achieves an average ATE of 0.43 cm and F1 of 88.0 using solely RGB images as input.}
			\label{fig:teaser}
		\end{center}
		\vspace{-3em}
	}

	%================================================
	% FOR RAL JOURNAL HEADERS
	%================================================
%	\markboth{ IEEE ROBOTICS AND AUTOMATION LETTERS. PREPRINT VERSION. ACCEPTED. June, 2024}{Naumann \MakeLowercase{\textit{et al.}}: NeRF-VO: Real-Time Sparse Visual Odometry with Neural Radiance
%		Fields} 
	%================================================
	%================================================
	
	\makeatletter
	\apptocmd{\@maketitle}{\centering\insertfig}{}{}
	\makeatother
	
	\maketitle

	\begin{abstract}
We introduce a novel monocular visual odometry (VO) system, NeRF-VO, that integrates learning-based sparse visual odometry for low-latency camera tracking and a neural radiance scene representation for fine-detailed dense reconstruction and novel view synthesis.
Our system initializes camera poses using sparse visual odometry and obtains view-dependent dense geometry priors from a monocular prediction network. We harmonize the scale of poses and dense geometry, treating them as supervisory cues to train a neural implicit scene representation.
NeRF-VO demonstrates exceptional performance in both photometric and geometric fidelity of the scene representation by jointly optimizing a sliding window of keyframed poses and the underlying dense geometry, which is accomplished through training the radiance field with volume rendering. 
We surpass SOTA methods in pose estimation accuracy, novel view synthesis fidelity, and dense reconstruction quality across a variety of synthetic and real-world datasets while achieving a higher camera tracking frequency and consuming less GPU memory.
\end{abstract} 
	\vspace{-2em}
	\section{Introduction}
\label{sec:introduction}

\begin{figure*}[t]
  \centering
  \includegraphics[width=\textwidth]{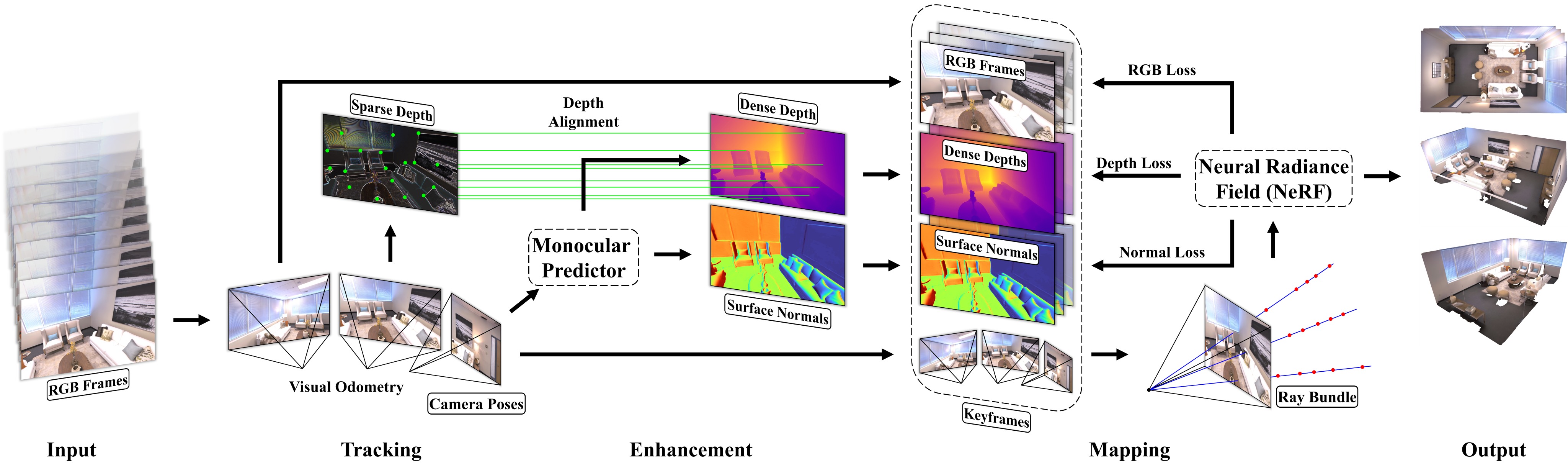}
  \caption{\textbf{System architecture of NeRF-VO.} The method uses only a sequence of RGB images as input. The sparse visual tracking module selects keyframes from this input stream and calculates camera poses and depth values for a set of sparse patches. Additionally, the dense geometry enhancement module predicts dense depth maps and surface normals and aligns them with the sparse depth from the tracking module. The NeRF-based dense mapping module utilizes raw RGB images, inferred depth maps, surface normals, and camera poses to optimize a neural implicit representation and refine the camera poses. Our system is capable of performing high-quality 3D dense reconstruction and rendering images at novel views.}
  \label{fig:system-design}
  \vspace{-1em}
\end{figure*}

\IEEEPARstart{A}{ccurate}  pose estimation and 3D reconstruction of the environment using images are fundamental challenges in 3D computer vision and essential prerequisites for diverse applications in robotics.
Neural Radiance Fields~(NeRF)~\cite{mildenhall2020nerf} have proven to be an excellent scene representation method for novel view synthesis tasks. The original NeRF employs large multi-layer perceptrons (MLP) to decode 3D coordinates and ray directions into volume density and color. However, it fails to represent complex and large scenes, particularly those with fine detail. Both training and rendering NeRFs with deep MLPs are computationally intensive and time-consuming.
Recent works~\cite{Chen2022ECCV, yu2022plenoxels, mueller2022instant} have sought to accelerate NeRFs by replacing deep MLPs with more efficient representations. For instance, Instant-NGP~\cite{mueller2022instant} achieves a substantial speedup by utilizing a hybrid representation that combines trainable multi-resolution hash encodings (MHE) with shared shallow MLPs. These accelerated neural implicit scene representations pave the way for real-time critical SLAM/VO.

NeRF-based scene representations enable high-fidelity photometric and geometric reconstructions while being very memory efficient.
They enable comprehensive exploitation of information from raw images, leveraging every pixel in the optimization process. This has the potential to push the performance frontier beyond traditional direct~\cite{engel2014lsd, engel2017direct, zubizarreta2020direct} and indirect SLAM~\cite{davison2007monoslam, campos2021orb, cvivsic2022soft2}.
Lately, numerous works have aimed at integrating SLAM with neural implicit mapping~\cite{Sucar:etal:ICCV2021, niceslam, pointslam, coslam, voxfusion, featurerealisticneural}. Yet, only a few focus on monocular RGB input~\cite{rosinol2022nerf, zhu2023nicerslam, li2023dense, orbeezslam, goslam, hislam}. In terms of accuracy, RGB-only methods lag behind their \mbox{RGB-D} counterparts, especially in dense reconstruction. Further, most of these approaches are computationally expensive, lack real-time capability, and require significant GPU memory.

To address these issues, we propose a real-time capable sparse visual odometry with neural implicit dense mapping. A preview of its performance is provided in Fig.~\ref{fig:teaser}. We obtain the initial pose estimation and sparse 3D landmarks using low-latency learning-based sparse visual odometry. Up-to-scale dense geometric cues, including monocular dense depth and normals, are inferred using a transformer-based neural network. 
With the initial poses, camera-captured monocular images, and dense geometric priors, we can efficiently optimize a neural radiance field that implicitly represents the 3D scene. 
Accurate poses and dense geometry of the scene are recovered by minimizing the disparity of captured images and predicted dense geometric cues relative to the renderings generated from the neural radiance field.
Hence, our proposed method comprises three main components: a sparse visual tracking front-end, a dense geometry enhancement module, and a NeRF-based dense mapping back-end. The system architecture is depicted in Fig.~\ref{fig:system-design}.

Overall, we introduce NeRF-VO, a neural SLAM system that employs sparse visual odometry for efficient pose estimation, and a NeRF scene representation for highly accurate dense mapping. It showcases superior geometric and photometric reconstruction in comparison to SOTA methods while maintaining the lowest tracking latency and GPU memory consumption among competing works. The main contributions of our work can be summarized as follows:
\begin{itemize}
\item We propose NeRF-VO, a monocular RGB SLAM system that utilizes sparse visual odometry for pose tracking and an implicit neural representation for mapping. This enables highly accurate camera tracking, promising 3D reconstruction, and high-fidelity novel view synthesis.
\item We present a novel paradigm for optimizing a NeRF scene representation and camera poses by incorporating dense depth and surface normal supervision. Utilizing a monocular transformer-based network, we predict dense depth and surface normal priors. To align their scale with the sparse visual odometry, we propose a dedicated sparse-to-dense scale alignment procedure.
\item NeRF-VO demonstrates SOTA performance in camera tracking, 3D dense reconstruction, and novel view synthesis across various synthetic and real-world datasets.
% \item We will open-source our versatile neural SLAM framework, fostering further research by allowing easy integration of other neural representations and VO/SLAM methods.
\end{itemize}
%
% In the remainder of this paper, we review related work in Sec.~\ref{sec:relatedwork} and introduce the sparse visual tracking front-end in Sec.~\ref{sec:visual_tracking}. We present the main methodology, which includes dense geometry enhancement, scale alignment, neural implicit mapping and dedicated NeRF optimization, in Sec.~\ref{sec:methodology}. We report and analyze extensive experimental results in Sec.~\ref{sec:experiments}. Finally, we conclude the paper and provide an overview of future work in Sec.~\ref{sec:conclusion}.
	\section{Related Work}
\label{sec:relatedwork}
For brevity, this section covers only the related work most relevant to our approach. Please refer to the survey by Tosi et al.\ \cite{survey} for a more detailed overview of recent advances in NeRF- and GS (Gaussian Splatting)-based SLAM.

\PAR{Learning-Based Visual Tracking.} Our work centers on monocular visual odometry using RGB image sequences from a calibrated camera. This technique estimates the camera positions and orientations of each incoming frame. Unlike SLAM, visual odometry focuses on local coherence among consecutive frames and does not include SLAM's loop closure optimization or global bundle adjustment. Recently, VO methods have advanced from traditional handcrafted feature detection and matching to deep learning-based approaches, enhancing accuracy and robustness~\cite{cnnslam, yang2020d3vo, wang2021tartanvo, teed2021droid, teed2022deep}. Among all of them, DROID-SLAM~\cite{teed2021droid} and DPVO~\cite{teed2022deep} are two noteworthy works that leverage neural networks to predict the optical flow between consecutive images and iteratively update camera poses. DPVO~\cite{teed2022deep} serves as the foundation for our system's front-end since it is highly efficient and accurate.

\PAR{Dense Visual SLAM.} Dense visual SLAM aims to construct a dense 3D representation of the environment instead of sparse 3D landmarks. Following the first real-time dense visual work DTAM~\cite{dtam}, many approaches have been proposed, primarily those exploiting monocular depth prediction~\cite{deepfactors, deeptam, demon}. The scene representations selected in these works have also progressed from volumetric representations~\cite{dtam} to low-dimension latent representations~\cite{codeslam, deepfactors, zuo2021codevio} and the integration of pre-trained depth estimators~\cite{cnnslam, tandem, xin2023simplemapping}. In this work, we choose a volumetric NeRF as the representation due to its superior photometric and geometric accuracy.

\PAR{NeRF-Enabled SLAM.} Recently, many works have been proposed to integrate NeRF-based~\cite{mildenhall2020nerf} scene representations into SLAM. In general, existing methods can be differentiated into \textit{one-stage} and \textit{two-stage} approaches.
\textit{Two-stage} approaches use an existing SLAM algorithm as a tracking module to estimate depth and camera poses, and then use these estimates as supervisory signals to optimize an implicit neural representation as part of a mapping module. 
Early approaches such as Orbeez-SLAM~\cite{orbeezslam} and NeRF-SLAM~\cite{rosinol2022nerf} demonstrated the effectiveness of this combination. Recent advances in this direction have further introduced view-centric implicit functions~\cite{NEWTON}, global loop closure~\cite{goslam, hislam}, and monocular depth priors~\cite{hislam} to improve pose estimation accuracy and dense mapping quality. 
In contrast, \textit{one-stage} approaches use a single implicit neural function for both tracking and mapping. iMAP~\cite{Sucar:etal:ICCV2021} and NICE-SLAM~\cite{niceslam} set the groundwork for this direction using RGB-D data. Subsequent works aimed at enhancing the scene representation~\cite{pointslam, coslam, voxfusion}, introducing implicit semantic encoding~\cite{featurerealisticneural}, and integrating inertial measurements~\cite{Lisus2023TowardsOW}. Among these, NICER-SLAM~\cite{zhu2023nicerslam} and DIM-SLAM~\cite{li2023dense} perform dense SLAM only using monocular images. NICER-SLAM prioritizes scene reconstruction and novel view synthesis through heavy optimization with various losses, while DIM-SLAM focuses on accurate camera tracking, but shows suboptimal performance in dense reconstruction. However, both methods are not suitable for real-time applications.

\PAR{GS-Based SLAM.} Recently, 3D Gaussian Splatting (3DGS), introduced by Kerbl et al.\ \cite{3DGS}, has attracted attention for its superior ability to model novel view synthesis using differentiable 3D Gaussian-shaped primitives. Recently, this novel paradigm has inspired a variety of SLAM approaches: Analogous to the NeRF-based \textit{one-stage} approaches, Gaussian Splatting SLAM~\cite{GSSLAM}, GS-SLAM~\cite{GS-SLAM}, and SplaTAM~\cite{SplaTAM} all track the camera using a dense photometric error, while Photo-SLAM~\cite{Photo-SLAM} uses ORB-SLAM3~\cite{campos2021orb} for tracking, similar to the NeRF-based \textit{two-stage} approaches. However, 3DGS does not explicitly model geometry, which hinders its use in robotic tasks that require geometric space awareness.

\PAR{Our work} follows a \textit{two-stage} design. We use a sparse visual tracking method, DPVO~\cite{teed2022deep} to get initial poses at high frequency and obtain dense depth and surface normals priors from a monocular prediction network. These initial poses and dense geometry cues are used to train our mapping back-end employing a nerfacto-based neural radiance field~\cite{nerfstudio}. Our system yields high-frequency pose estimations and promising dense reconstruction after neural scene optimization, all achieved at a low memory footprint.
	\section{Sparse Visual Pose Tracking}\label{sec:visual_tracking}

We employ the Deep Patch Visual Odometry (DPVO)~\cite{teed2022deep} algorithm as our tracking front-end. DPVO is a sparse, monocular, learning-based algorithm that estimates camera poses and sparse depths for a set of patches per keyframe.

\PAR{Patch Graph.} Given a sequence of RGB frames, DPVO randomly samples a set of $ K $ square patches of size $ s \times s$ pixels per keyframe and adds them to a bipartite patch graph that connects patches and frames. For instance, the $ k $-th square patch from frame $ i $ represented by $ \mathbf{P}^{i}_{k}=\begin{bmatrix} \mathbf{u}^T & \mathbf{v}^T & \mathbf{1}^T & \mathbf{d}^T \end{bmatrix}^T $ is connected via edges to all frames within temporal vicinity of frame $ i $. $ \mathbf{u}, \mathbf{v} $ represent the pixel coordinates and $ \mathbf{d} $ the inverse depths with $ \mathbf{u}, \mathbf{v}, \mathbf{d} \in \mathbb{R}^{1 \times s^2} $. $ \mathbf{1} $ is a vector filled with ones. The patch graph builds a trajectory for each patch, incorporating all of its reprojections $ \mathbf{r}^{ij}_{k} $, where $ j \in \mathcal{N} $, and $ \mathcal{N} $ is the array of frames temporally adjacent to $ i $. Assuming uniform depth across each patch, the reprojection is defined by:
\begin{equation}
\mathbf{r}^{ij}_{k} \sim \mathbf{K} \mathbf{T}_{j} \mathbf{T}_{i}^{-1} \mathbf{K}^{-1} \mathbf{P}^i_{k} \text{,}
\end{equation}
where $ \mathbf{K} $ is the $4 \times 4$ camera calibration matrix, and $ \mathbf{T}_{i} \in \mathbb{SE}(3) $ represents the world-to-camera transformation of frame $ i $. For simplicity, we omit the normalization operation of the third dimension in the above equation. 

\PAR{Differentiable Bundle Adjustment.} The key component of DPVO is its differentiable pose and depth update operator. A recurrent neural network operates on the patch graph with a set of edges $ \mathcal{E} $, while maintaining a hidden state for each edge $ (k,i,j) \in \mathcal{E} $ (patch-frame pair), and predicting a 2D correction vector $ \pmb{\delta} ^{ij}_k \in \mathbb{R}^2 $ for the reprojection of the patch center $\bar{\mathbf{r}}^{ij}_{k}$ with a corresponding confidence weight $ \pmb{\psi}^{ij}_k \in \mathbb{R}^2 $.
Bundle adjustment is performed using the optical flow correction as a constraint to iteratively update frame poses and patch depths via nonlinear least-squares optimization. The cost function for bundle adjustment is:
\begin{equation}
\sum_{(k,i,j) \in \mathcal{E}} {\left\lVert \mathbf{K} \mathbf{T}_{j} \mathbf{T}_{i}^{-1} \mathbf{K}^{-1} \bar{\mathbf{P}}_{k}^{i} - \left[ \bar{\mathbf{r}}^{ij}_{k} + \pmb \delta^{ij}_k \right] \right\rVert}^2_{\pmb \psi^{ij}_k} \text{,}
\end{equation}
where $ {\left\lVert \cdot \right\rVert}_{\pmb \psi} $ denotes the Mahalanobis distance. The bundle adjustment step of DPVO is differentiable, the recurrent neural network is trained on a set of ground truth poses, together with the corresponding ground truth optical flow to supervise pose estimation and optical flow correction. 

\PAR{Keyframing Strategy.} DPVO optimizes the keyframes' patch depths and camera poses in a sliding optimization window, adding each incoming frame. This sliding window keeps a constant array of the most recent keyframes and removes old ones to maintain bounded computational complexity ensuring runtime efficiency. The three most recent frames are always considered to be keyframes, but the fourth most recent keyframe is only kept if the optical flow to its predecessor is sufficiently high, otherwise, it is removed.
We use this keyframe selection for our downstream geometry enhancement and mapping modules. To stabilize NeRF optimization, the last three keyframes are not propagated yet, as they are likely to be removed later by the keyframing scheme described above. Each time the fourth most recent keyframe is secured, it is passed to our dense geometry enhancement module and NeRF optimization, including all camera poses and patches in the current sliding window.
Since only certain keyframes are involved in our dense geometry enhancement and NeRF optimization, the computational and memory footprint of NeRF-VO are bounded. 

\section{Neural Implicit Dense Mapping}
\label{sec:methodology}

As indicated in the system overview in Fig.~\ref{fig:system-design}, the keyframes from the visual tracking front-end are subsequently processed by our dense geometry enhancement and incorporated into our NeRF optimization back-end.

\subsection{Dense Geometry Enhancement}

Since the sparse patch depths $ \mathbf{D}_{s} $ from the visual tracking module cover only a tiny portion of the image, we incorporate a dense geometry enhancement module that predicts a dense depth map $ \mathbf{D}_{d} $ based solely on the monocular RGB input. We then align this dense depth prediction to the sparse patch depths using a dedicated scale alignment procedure. Monocular depth prediction is performed using the off-the-shelf network \cite{eftekhar2021omnidata} based on the Dense Prediction Transformer (DPT) architecture \cite{Ranftl2020, Ranftl2021}.

\PAR{Scale Alignment.} Since both sparse and dense depths are predicted solely based on RGB input, their depth distributions are skewed relative to the metric depth and each other. However, downstream dense mapping necessitates depth maps and camera poses at a unified scale. Therefore, we introduce a dense-to-sparse scale alignment procedure to align the predicted dense depth map of each keyframe to the sparse patch depth. Assuming that the sparse depth $ \mathbf{D}_{s} $ and the dense depth $ \mathbf{D}_{d} $ follow a Gaussian distribution with $ \mathbf{D}_{s} \sim \mathcal{N}(\mu_{{s}},\sigma^{2}_{{s}}) $ and $ \mathbf{D}_{d} \sim \mathcal{N}(\mu_{{d}},\sigma^{2}_{{d}}) $, we compute the aligned dense depth by:
\begin{equation}
\mathbf{D}'_{d} = \alpha \mathbf{D}_{d} + \beta \mathbf{1} \text{,}
\end{equation}
where $ \alpha $ and $ \beta $ are a scale and shift value defined as: 
\begin{equation}
\label{eq:alignment}
\alpha = \frac{\sigma_{ {s}}}{\hat{\sigma}_{{d}}} \text{, } \beta = \mu_{{d}} \left( \frac{\mu_{{s}}}{\hat{\mu}_{{d}}} - \alpha \right) \text{,}
\end{equation}
where $ \hat{\mu}_{{d}} $ and $ \hat{\sigma}_{{d}} $ are the mean and standard deviation of the sparsified dense depth map $ \hat{\mathbf{D}}_{d} $ extracted from ${\mathbf{D}}_{d} $ at the pixel coordinates of $ \mathbf{D}_{s} $. Since the patches are sampled randomly, given a sufficiently large sample size per frame, one could assume that $\mu_{{d}} \approx \hat{\mu}_{{d}}$. This would lead to a relaxed formulation of $ \beta $ with $ \beta = \mu_{s} - \alpha\hat{\mu}_{{d}} $. We have empirically evaluated this relaxed alignment scheme along with several other alignment strategies and found that the standard formulation in Eq.~\ref{eq:alignment} performs best (see Sec.~\ref{ablations}).

\PAR{Surface Normal Cues.} To improve the accuracy of the neural implicit representation for fine-grained dense mapping, we introduce surface normal constraints by matching against surface normals predicted from the RGB input using the same off-the-shelf monocular predictor from \cite{eftekhar2021omnidata}.

\subsection{Mapping with Neural Scene Representation}
\PAR{Neural Radiance Field.} Our dense mapping back-end reconstructs the scene by optimizing a neural radiance field built on the nerfacto model~\cite{nerfstudio}. The neural scene representation comprises multi-resolution hash encodings similar to~\cite{mueller2022instant}, spherical harmonics encoding, and two multi-layer perceptrons (MLPs) for density and color recovery. The NeRF scene representation is optimized using the camera poses, captured RGB images as well as inferred depth maps and surface normals of all keyframes.

We generate ray bundles $ \mathcal{R} $ inside the camera frustums of the keyframes and sample 3D coordinates along these rays $ t \in \mathcal{R} $ by using a proposal sampler, which is optimized jointly with the rest of the architecture. Each sampled 3D coordinate is mapped to a set of learnable embeddings in the multi-resolution hashmap. These embeddings are processed by the first MLP to predict the volume density. The resulting density is fed into the second MLP together with the spherical harmonic encoded view direction and appearance embeddings to compute the color. The surface normals per sample are obtained by computing the gradient over the first MLP with respect to the hashed 3D coordinate. Finally, the color, depth, and surface normal for the entire ray are obtained using volumetric rendering similar to \cite{mildenhall2020nerf}. 

\PAR{Joint Optimization of Poses and Neural Scene.}
Since the keyframe camera poses $ \mathbf{T}_C $ are involved in ray sampling, they can be optimized jointly with the NeRF scene representation. Hence, we optimize the learnable parameters $ \pmb \Theta $ of the scene representation and the MLPs jointly with the camera poses $ \mathbf{T}_C $ using a weighted sum of four losses.

Our color loss is defined as the mean squared error between the rendered color image and the ground truth RGB input, analogous to \cite{mildenhall2020nerf}:
\begin{equation}
{L}_{\rm{rgb}}\left( \mathbf{T}_C, \pmb \Theta\right) = \sum_{\left( u,v \right) \in \mathcal{B}} \left\lVert \mathbf{C}\left[ u,v \right] - \breve{\mathbf{C}}\left[ u,v \right] \left( \mathbf{T}_C, \pmb \Theta\right) \right\rVert_2^2 \text{,}
\end{equation}
where $ \breve{\mathbf{C}}\left[ u,v \right] $ and $ \mathbf{C}\left[ u,v \right] $ are the rendered and ground truth color, and $ \left( u, v \right) $ are the pixel coordinates of the ray bundle.

To counter noise and bias in our depth estimation and alignment procedures, we employ the uncertainty-aware depth loss~\cite{depthloss}, which aims to minimize the Kullback-Leibler divergence between the assumed Gaussian distributed depth values of all samples along each ray and our aligned dense depth map $ \mathbf{D}'_{d} $:
\begin{multline}
{L}_{d}\left( \mathbf{T}_C, \pmb \Theta\right) = \sum_{\left( u, v \right) \in \mathcal{B}} \sum_{t \in \mathcal{R}\left[ u,v \right] } \log{\left(\mathcal{T}_t\left( \mathbf{T}_C, \pmb \Theta\right)\right)} \cdot \\ 
\exp{\left(-\frac{\left( d_t - \mathbf{D}'_{d}\left[ u,v \right] \right)^2}{2\hat{\sigma}^2}\right)} \Delta d_t \text{,}
\end{multline}
where $t \in \mathcal{R}\left[ u,v \right] $ is the $ t $-th sample along the ray $ \mathcal{R}\left[ u,v \right] $, $ d_t $ is the distance of the sample $ t $ from the camera center along ray $ \mathcal{R}\left[ u,v \right] $, $ \Delta d_t $ is the distance between the sampling distances $ d_{t+1} $ and $ d_t $ with $ \Delta d_t = d_{t+1} - d_t $, $ \hat{\sigma} $ is the estimated variance of the depth $ \mathbf{D}'_{d}\left[ u,v \right] $ (we use $ 0.001 $ empirically) and $ \mathcal{T}_t\left( \mathbf{T}_C, \Theta\right) $ is the accumulated transmittance along the ray $ \mathcal{R}\left[ u,v \right] $ up to sample $ t $ computed by:
\begin{equation}
\mathcal{T}_t\left( \mathbf{T}_C, \pmb \Theta\right) = \exp{\left( -\sum_{s < t}\rho_s\left( \mathbf{T}_C, \pmb \Theta\right) \Delta d_s \right)} \text{,}
\end{equation}
where $ \rho_s\left( \cdot \right) $ is the estimated density at the sample $ s $.

The normal supervision is guided by the normal consistency loss proposed in \cite{normalloss}, which integrates the L1 distance and cosine distance between the rendered surface normals $ \breve{\mathbf{N}}\left[ u,v \right] \in \mathbb{R}^3 $ and the normals predicted from the monocular prediction network $ \mathbf{N}\left[ u,v \right] $ at pixel $ \left( u, v \right) $:
\begin{multline}
{L}_{n}\left( \mathbf{T}_C, \pmb \Theta\right) = \sum_{\left( u,v \right) \in \mathcal{B}} \left\lVert \mathbf{N}\left[ u,v \right] - \breve{\mathbf{N}}\left[ u,v \right]\left( \mathbf{T}_C, \pmb \Theta\right) \right\rVert_1 + \\
\left\lVert 1 - \mathbf{N}\left[ u,v \right] ^\top \breve{\mathbf{N}}\left[ u,v \right]\left( \mathbf{T}_C, \pmb \Theta\right) \right\rVert_1 \text{,}
\end{multline}
where $ \mathbf{N}\left[ u,v \right] ^\top \breve{\mathbf{N}}\left[ u,v \right] $ indicates the cosine similarity.

Finally, we use two regularization terms for the volume density. The distortion and proposal losses introduced by \cite{interlevelloss} serve to prevent floaters and background collapse and guide the optimization of the proposal sampler:
\begin{equation}
{L}_{\rm{reg}} = {L}_{\rm{prop}} + 0.002 {L}_{\rm{dist}} \text{,}
\end{equation}

Overall, the complete loss for our NeRF optimization is:
\begin{equation}
{L} = {L}_{\rm{rgb}} + 0.001 {L}_{d} + 0.00001 {L}_{n} + {L}_{\rm{reg}}
\end{equation}

\subsection{System Design}
The sparse tracking, dense geometry enhancement, and dense mapping modules run asynchronously on separate threads. Information is propagated in a feed-forward fashion through the three components sequentially. Communication occurs each time a new keyframe is secured in the visual tracking module. It is then processed by the enhancement module and included in the training database of the mapping module. Separated from this process, the mapping module continuously optimizes the implicit representation jointly with the camera poses. With this multi-threaded architecture, our system can run in real-time, with the visual tracking running at low latency, while the enhancement and mapping run at a relatively low frequency, which does not block the data streaming in from the sensor. Due to its asynchronous architecture, our system has the potential to be run on multiple GPUs, leading to improved runtime efficiency.
	\section{Experiments}
\label{sec:experiments}
We evaluate our method qualitatively and quantitatively against the state-of-the-art monocular SLAM methods using neural implicit representations on a variety of real and synthetic indoor datasets and present an extensive ablation study of the key design choices of our method.

\subsection{Evaluation Datasets}
We evaluate our method on four datasets. Replica~\cite{replica19arxiv}: a synthetic dataset that contains high-quality reconstructions of indoor scenes, where we use the eight trajectories generated by~\cite{Sucar:etal:ICCV2021}. ScanNet~\cite{dai2017scannet}: a real-world dataset where the \mbox{RGB-D} sequence is captured using an RGB-D sensor attached to a handheld device and the ground truth trajectory is computed using BundleFusion~\cite{dai2017bundlefusion}. 7-Scenes~\cite{shotton2013scene}: a real-world dataset that provides a trajectory of RGB-D frames captured by Kinect sensors where the ground truth trajectories and dense 3D models are computed using KinectFusion~\cite{kinectfusion}.
We also evaluate our method on a self-captured dataset consisting of a set of four RGB-D trajectories captured with the RealSense sensor in a meeting room. The ground-truth trajectories are computed using the visual-inertial SLAM system OKVIS2~\cite{leutenegger2022okvis2} based on synchronized stereo images and IMU data, while the 3D model is obtained using TSDF-Fusion~\cite{kinectfusion} on the captured depth maps.

\subsection{Baselines}
We compare our method against the SOTA SLAM method NeRF-SLAM~\cite{rosinol2022nerf}, which also uses neural implicit representations for mapping. Additionally, where applicable, we report results from the contemporary methods DIM-SLAM~\cite{li2023dense}, NICER-SLAM~\cite{zhu2023nicerslam}, Orbeez-SLAM~\cite{orbeezslam}, GO-SLAM~\cite{goslam} and HI-SLAM~\cite{hislam}. To evaluate the benefits of our back-end, we compare the camera tracking accuracy against DPVO \cite{teed2022deep} as well as the visual odometry version of DROID-SLAM~\cite{teed2021droid} (DROID-VO), which is used as the tracking module in NeRF-SLAM.

\subsection{Metrics}
We evaluate our method on three tasks: camera tracking, novel view synthesis, and 3D dense reconstruction. For camera tracking, we align the estimated trajectory to the ground truth using the Kabsch-Umeyama~\cite{Kabsch:a15629, 88573} algorithm and evaluate the accuracy using the absolute trajectory error (ATE RMSE)~\cite{sturm12iros}. For novel view synthesis, we assess the quality of the rendered RGB images using peak signal-to-noise ratio (PSNR), structural similarity (SSIM)~\cite{ssim}, and learned perceptual image patch similarity (LPIPS)~\cite{lpips}. To compute these metrics, we select 125 equally spaced RGB frames along the trajectory. To render the exact view of these evaluation frames, we transform the ground truth camera poses to the coordinate system of the implicit function. For 3D reconstruction, we consider accuracy, completion, and recall. We render the predicted mesh from our neural implicit representation. Analogous to \cite{rosinol2022nerf, li2023dense, zhu2023nicerslam, goslam, hislam} we exclude regions not observed by any camera from the evaluation. We align both meshes using the Iterative Closest Point (ICP) algorithm. To ensure statistical validity, we report the average of 5 independent runs.

\subsection{Camera Tracking}
Table~\ref{tab:cam_replica} reports the camera tracking results on the Replica dataset~\cite{replica19arxiv, Sucar:etal:ICCV2021}, showing that our model achieves the lowest average ATE RMSE over all scenes. The results of DIM-SLAM~\cite{li2023dense} and NICER-SLAM~\cite{zhu2023nicerslam} are taken from their original papers. While DIM-SLAM~\cite{li2023dense} and NeRF-SLAM~\cite{rosinol2022nerf} perform better on some scenes, their performance deteriorates on others, whereas our method remains relatively stable across all scenes.
Table~\ref{tab:cam_scannet} reports the tracking accuracy on six commonly tested scenes from ScanNet~\cite{dai2017scannet}. 
Our method outperforms concurrent visual odometry methods whose tracking module only applies implicit ``local loop closure" (L-LC) imposed by the scene representations (maps), such as HI-SLAM (VO), NICER-SLAM, DROID-VO, NeRF-SLAM, and DPVO. In general, we can conclude that the full SLAM methods incorporating global loop closure (G-LC) and scene-wide global bundle adjustment perform better on these large scenes.
Finally, Table~\ref{tab:cam_custom} shows the performance of our method and NeRF-SLAM~\cite{rosinol2022nerf} as well as their respective tracking modules on our custom dataset, demonstrating the superior camera tracking performance of our model.
{
\def\arraystretch{1.0}
\setlength\tabcolsep{3.5pt}
\setlength\arrayrulewidth{0.5pt}
\begin{table}[tbp]
\scriptsize
\caption{Camera tracking performance (ATE RMSE [cm]) on the Replica~\cite{replica19arxiv} dataset.}
\vspace{-0.5em}
\centering
\begin{tabular}{l|c c c c c c c c|c}
\toprule
\textbf{Model} & {\textbf{rm-0}} & {\textbf{rm-1}} & {\textbf{rm-2}} & {\textbf{of-0}} & {\textbf{of-1}} & {\textbf{of-2}} & {\textbf{of-3}} & {\textbf{of-4}} & {\textbf{Avg.}} \\
\midrule
DIM-SLAM \cite{li2023dense} & 0.48 & 0.78 & 0.35 & 0.67 & 0.37 & \textbf{0.36} & \textbf{0.33} & \textbf{0.36} & 0.46 \\
NICER-SLAM \cite{zhu2023nicerslam} & 1.36 & 1.60 & 1.14 & 2.12 & 3.23 & 2.12 & 1.42 & 2.01 & 1.88 \\
\midrule
DROID-VO \cite{teed2021droid} & 0.50 & 0.70 & 0.30 & 0.98 & \textbf{0.29} & 0.84 & 0.45 & 1.53 & 0.70 \\
NeRF-SLAM \cite{rosinol2022nerf} & 0.40 & 0.61 & \textbf{0.20} & \textbf{0.21} & 0.45 & 0.59 & \textbf{0.33} & 1.30 & 0.51 \\
DPVO \cite{teed2022deep} & 0.49 & 0.54 & 0.54 & 0.77 & 0.36 & 0.57 & 0.46 & 0.57 & 0.54 \\
\textbf{Ours} & \textbf{0.28} & \textbf{0.33} & 0.36 & 0.53 & 0.40 & 0.53 & 0.56 & 0.47 & \textbf{0.43} \\
\bottomrule
\end{tabular}
%\vspace{-1em}
\label{tab:cam_replica}
\end{table}
}
{
\def\arraystretch{1.0}
\setlength\tabcolsep{4.6pt}
\setlength\arrayrulewidth{0.5pt}
\begin{table}[tbp]
\scriptsize
\caption{Camera tracking performance (ATE RMSE [cm]) on ScanNet~\cite{dai2017scannet}.
The methods are categorized according to whether they perform global (G) or only local (L) loop closure (LC) optimization.}
\vspace{-0.5em}
\centering
\begin{tabular}{l|l|c c c c c c|c}
\toprule
& \textbf{Model} & \textbf{0000} & \textbf{0059} & \textbf{0106} & \textbf{0169} & \textbf{0181} & \textbf{0207} & \textbf{Avg.} \\
\midrule
\multirow{3}{*}{\rotatebox[origin=c]{90}{G-LC}} & {Orbeez-SLAM} \cite{orbeezslam} & 7.2 & \textbf{7.2} & 8.1 & \textbf{6.6} & 15.8 & \textbf{7.2} & 8.7 \\
& {GO-SLAM} \cite{goslam} & \textbf{5.9} & 8.3 & 8.1 & 8.4 & 8.3 & n/a & n/a \\
& {HI-SLAM} \cite{hislam} & 6.4 & \textbf{7.2} & \textbf{6.5} & 8.5 & \textbf{7.6} & 8.4 & \textbf{7.4} \\
\midrule
\multirow{5}{*}{\rotatebox[origin=c]{90}{L-LC}} & {HI-SLAM (VO)} \cite{hislam} & 14.4 & 26.7 & \textbf{10.0} & 15.5 & 9.3 & 9.9 & 14.3 \\
\cmidrule{2-9}
& {NICER-SLAM} \cite{zhu2023nicerslam} & 96.0 & 63.2 & 117.1 & 77.9 & 68.0 & 70.8 & 82.2 \\
& {DROID-VO} \cite{teed2021droid} & 14.4 & \textbf{16.4} & 10.9 & 16.3 & 10.8 & 13.6 & 13.7\\
& {NeRF-SLAM} \cite{rosinol2022nerf} & 14.9 & 16.6 & 10.7 & 16.5 & 12.8 & 13.8 & 14.2 \\
& {DPVO} \cite{teed2022deep} & 13.1 & 19.2 & 12.4 & 13.3 & \textbf{8.8} & 8.9 & 12.6 \\
& \textbf{Ours} & \textbf{12.7} & 19.0 & 12.4 & \textbf{13.2} & 9.0 & \textbf{8.6} & \textbf{12.5} \\
\bottomrule
\end{tabular}
\vspace{-1em}
\label{tab:cam_scannet}
\end{table}
}
{
\setlength\tabcolsep{9.8pt}
\setlength\arrayrulewidth{0.5pt}
\begin{table}[tbp]
\scriptsize
\caption{Camera tracking performance (ATE RMSE [cm]) on our custom dataset.
}
\vspace{-0.5em}
\centering
\begin{tabular}{l|c c c c|c}
\toprule
\textbf{Model} & {\textbf{seq-1}} & {\textbf{seq-2}} & {\textbf{seq-3}} & {\textbf{seq-4}} & {\textbf{Avg.}} \\
\midrule
{DROID-VO} \cite{teed2021droid} & 2.58 & 1.37 & 1.81 & 3.05 & 2.20 \\
{NeRF-SLAM} \cite{rosinol2022nerf} & 2.32 & 1.19 & \textbf{1.31} & 5.26 & 2.52 \\
{DPVO} \cite{teed2022deep} & 2.96 & 1.38 & 1.57 & 3.14 & 2.26 \\
\textbf{Ours} & \textbf{2.08} & \textbf{1.07} & 1.43 & \textbf{2.68} & \textbf{1.81} \\
\bottomrule
\end{tabular}
\label{tab:cam_custom}
\end{table}
}

\subsection{3D Reconstruction}\label{sec:3d_reconstruction}
\begin{figure}[t]
  \centering
  \includegraphics[width=\columnwidth]{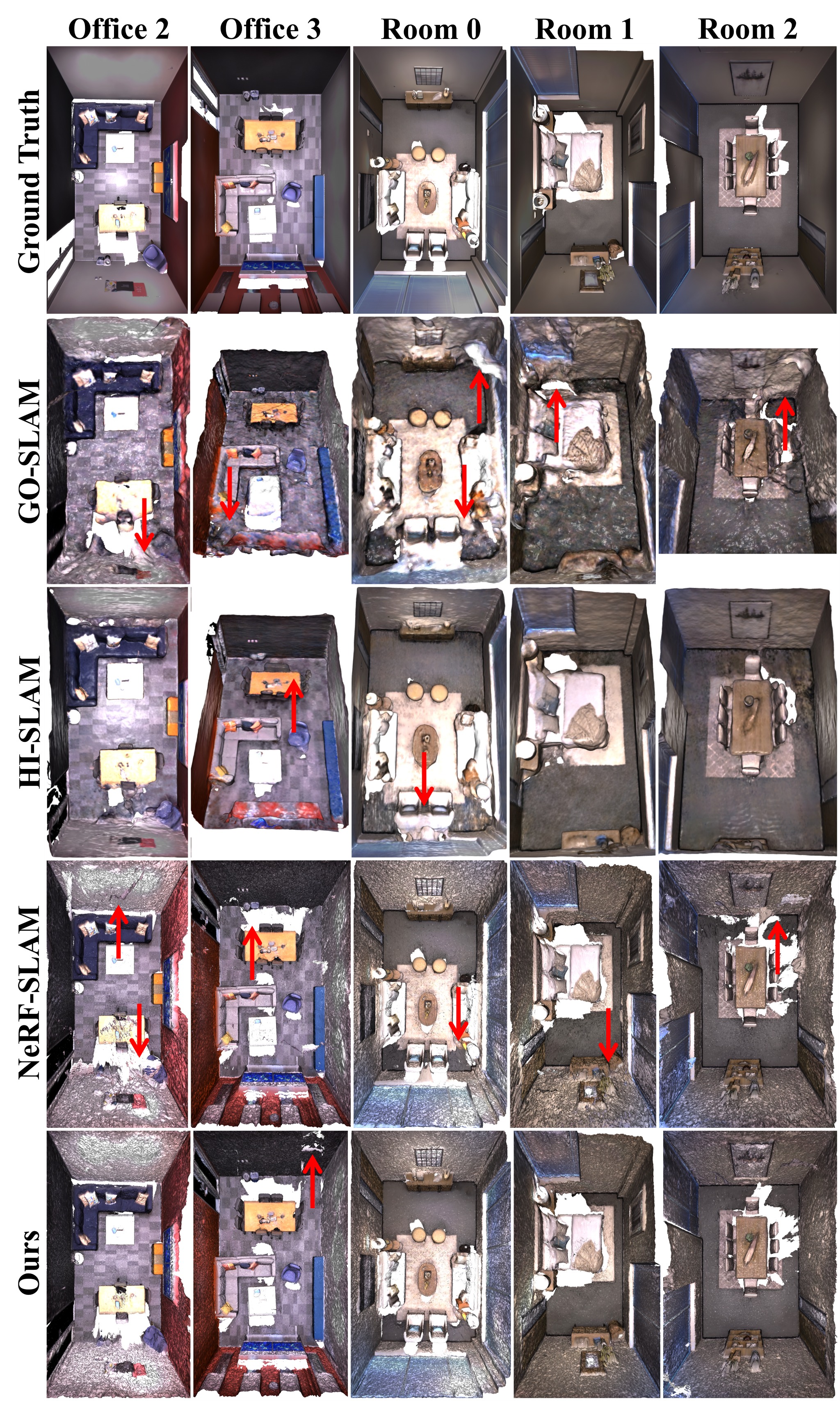}
  \caption{\textbf{3D reconstructions of five scenes from Replica~\cite{replica19arxiv}.} The pictures of GO-SLAM~\cite{goslam} and HI-SLAM~\cite{hislam} have been taken from their respective papers. Arrows highlight selected prominent artifacts and defects.}
  \label{fig:3dreconstruction}
  \vspace{-1em}
\end{figure}

Table~\ref{tab:rec_replica} demonstrates the superior 3D reconstruction performance of our model over all compared concurrent works on the Replica~\cite{replica19arxiv} dataset. Our method outperforms SOTA in accuracy, completion, and recall on four scenes and on average. The inferior performance on the office0 and office1 scenes is the result of scene-specific characteristics that complicate monocular depth predictions: these scenes have detail-rich wall paintings that lead to artifacts in monocular depth estimation. We notice that this is also the case for NICER-SLAM~\cite{zhu2023nicerslam}, which similarly relies on a monocular depth prediction network. A qualitative comparison of the 3D reconstruction is provided in Fig.~\ref{fig:3dreconstruction}.
Table~\ref{tab:rel_recon} reports the averaged 3D reconstruction performance of our method and NeRF-SLAM~\cite{rosinol2022nerf} on three real-world datasets: 7-Scenes~\cite{shotton2013scene}, ScanNet~\cite{dai2017scannet}, and our custom dataset. The results show that our model outperforms NeRF-SLAM on average on all datasets. Unfortunately, none of the concurrent papers report 3D reconstruction performance on any dataset other than the synthetic Replica~\cite{replica19arxiv}, thus we are unable to benchmark our model against other methods besides NeRF-SLAM.
{
\def\arraystretch{1.0}
\setlength\tabcolsep{3.6pt}
\setlength\arrayrulewidth{0.5pt}
\begin{table}[tbp]
\tiny
\caption{3D reconstruction results on Replica~\cite{replica19arxiv}.}
\vspace{-0.5em}
\centering
\begin{tabular}{l|l|c c c c c c c c|c}
\toprule
\textbf{Model} & \textbf{Metric} & {\textbf{rm-0}} & {\textbf{rm-1}} & {\textbf{rm-2}} & {\textbf{of-0}} & {\textbf{of-1}} & {\textbf{of-2}} & {\textbf{of-3}} & {\textbf{of-4}} & {\textbf{Avg.}} \\
\midrule
\multirow{3}{*}{{DIM-SLAM} \cite{li2023dense}} & {Acc.} [cm] \textdownarrow & 3.68 & 3.64 & 5.84 & 2.60 & \textbf{2.02} & 4.50 & 5.43 & 4.57 & 4.03 \\
& {Com.} [cm] \textdownarrow & 5.32 & 4.72 & 5.70 & \textbf{2.65} & 3.31 & 6.09 & 5.98 & 5.81 & 4.20 \\
& {Rec.} [\textless 5cm \%] \textuparrow & 82.2 & 80.4 & 74.44 & \textbf{89.6} & 84.9 & 75.3 & 73.5 & 76.6 & 79.6 \\
\midrule
\multirow{3}{*}{{NICER-SLAM} \cite{zhu2023nicerslam}} & {Acc.} [cm] \textdownarrow & 2.53 & 3.93 & 3.40 & 5.49 & 3.45 & 4.02 & 3.34 & 3.03 & 3.65 \\
& {Com.} [cm] \textdownarrow & 3.04 & 4.10 & 3.42 & 6.09 & 4.42 & 4.29 & 4.03 & 3.87 & 4.16 \\
& {Rec.} [\textless 5cm \%] \textuparrow & 88.8 & 76.6 & \textbf{86.1} & 65.2 & 77.8 & 74.5 & 82.0 & 84.0 & 79.4 \\
\midrule
\multirow{3}{*}{{GO-SLAM} \cite{goslam}} & {Acc.} [cm] \textdownarrow & 4.60 & 3.31 & 3.97 & 3.05 & 2.74 & 4.61 & 4.32 & 3.91 & 3.81 \\
& {Com.} [cm] \textdownarrow & 5.56 & 3.48 & 6.90 & 3.31 & 3.46 & 5.16 & 5.40 & 5.01 & 4.79 \\
& {Rec.} [\textless 5cm \%] \textuparrow & 73.4 & 82.9 & 74.2 & 82.6 & \textbf{86.2} & 75.8 & 72.6 & 76.6 & 78.0 \\
\midrule
\multirow{3}{*}{{HI-SLAM} \cite{hislam}} & {Acc.} [cm] \textdownarrow & 3.33 & 3.50 & 3.11 & 3.77 & 2.46 & 4.86 & 3.92 & 3.53 & 3.56 \\
& {Com.} [cm] \textdownarrow & 3.29 & 3.20 & \textbf{3.39} & 3.65 & 3.61 & 3.68 & 4.13 & 3.82 & 3.60 \\
& {Rec.} [\textless 5cm \%] \textuparrow & 86.4 & 85.8 & 83.0 & 80.7 & 82.4 & 82.9 & 80.3 & 82.3 & 83.0 \\
\midrule
\midrule
\multirow{3}{*}{{NeRF-SLAM} \cite{rosinol2022nerf}} & {Acc.} [cm] \textdownarrow & 2.77 & 4.50 & 3.45 & \textbf{1.88} & 2.09 & 3.77 & 3.24 & 3.06 & 3.10 \\
& {Com.} [cm] \textdownarrow & 3.45 & 3.49 & 6.32 & 3.76 & 3.19 & 4.20 & 4.23 & 3.97 & 4.08 \\
& {Rec.} [\textless 5cm \%] \textuparrow & 90.2 & 83.1 & 82.3 & 88.4 & 85.7 & 82.4 & \textbf{87.0} & 87.1 & 85.8 \\
\midrule
\multirow{3}{*}{\textbf{Ours}} & {Acc.} [cm] \textdownarrow & \textbf{2.24} & \textbf{1.89} & \textbf{3.02} & 3.45 & 3.15 & \textbf{3.30} & \textbf{3.05} & \textbf{2.35} & \textbf{2.81} \\
& {Com.} [cm] \textdownarrow & \textbf{2.96} & \textbf{2.33} & 6.04 & 3.83 & \textbf{3.15} & \textbf{3.41} & \textbf{3.64} & \textbf{3.36} & \textbf{3.59} \\
& {Rec.} [\textless 5cm \%] \textuparrow & \textbf{91.6} & \textbf{93.0} & 81.5 & 79.2 & 83.5 & \textbf{86.4} & 86.1 & \textbf{88.7} & \textbf{86.3} \\
\bottomrule
\end{tabular}
\vspace{-1em}
\label{tab:rec_replica}
\end{table}
}
{
\def\arraystretch{1.0}
\setlength\tabcolsep{6.6pt}
\setlength\arrayrulewidth{0.5pt}
\begin{table}[tbp]
\scriptsize
\caption{Averaged 3D reconstruction results on the real-world datasets: 7-Scenes~\cite{shotton2013scene}, ScanNet~\cite{dai2017scannet}, and our own.} 
\vspace{-0.5em}
\centering
\begin{tabular}{l|l|c c c c c c c|c}
\toprule
\textbf{Model} & \textbf{Metric} & {\textbf{7-Scenes}} & {\textbf{ScanNet}} & {\textbf{Custom}} \\
\midrule
\multirow{3}{*}{{NeRF-SLAM} \cite{rosinol2022nerf}} & {Acc.} [cm] \textdownarrow & 20.91 & 19.2 & 13.09 \\
& {Com.} [cm] \textdownarrow & 26.81 & 25.6 & 10.16 \\
& {Rec.} [\textless 5cm \%] \textuparrow & 28.5 & 21.1 & 47.6 \\
\midrule
\multirow{3}{*}{\textbf{Ours}} & {Acc.} [cm] \textdownarrow & \textbf{20.11} & \textbf{17.3} & \textbf{4.73} \\
& {Com.} [cm] \textdownarrow & \textbf{21.62} & \textbf{23.1} & \textbf{6.75} \\
& {Rec.} [\textless 5cm \%] \textuparrow & \textbf{34.0} & \textbf{22.5} & \textbf{65.0} \\
\bottomrule
\end{tabular}
\vspace{-1em}
\label{tab:rel_recon}
\end{table}
}

{
\def\arraystretch{1.0}
\setlength\tabcolsep{3.9pt}
\setlength\arrayrulewidth{0.5pt}
\begin{table}[tbp]
\tiny
\caption{Novel view synthesis results on Replica~\cite{replica19arxiv, Sucar:etal:ICCV2021}.
}
\vspace{-0.5em}
\centering
\begin{tabular}{l|l|c c c c c c c c|c}
\toprule
\textbf{Model} & \textbf{Metric} & {\textbf{rm-0}} & {\textbf{rm-1}} & {\textbf{rm-2}} & {\textbf{of-0}} & {\textbf{of-1}} & {\textbf{of-2}} & {\textbf{of-3}} & {\textbf{of-4}} & {\textbf{Avg.}} \\
\midrule
\multirow{3}{*}{{NICER-SLAM} \cite{zhu2023nicerslam}} & {PSNR} \textuparrow & 25.33 & 23.92 & 26.12 & 28.54 & 25.86 & 21.95 & 26.13 & 25.47 & 25.41 \\
& {SSIM} \textuparrow & 0.751 & \textbf{0.771} & \textbf{0.831} & 0.866 & 0.852 & \textbf{0.820} & \textbf{0.856} & \textbf{0.865} & \textbf{0.827} \\
& {LPIPS} \textdownarrow & 0.250 & 0.215 & 0.176 & 0.172 & 0.178 & 0.195 & 0.162 & 0.177 & 0.191 \\
\midrule
\multirow{3}{*}{{NeRF-SLAM} \cite{rosinol2022nerf}} & {PSNR} \textuparrow & 34.07 & 34.12 & \textbf{37.06} & \textbf{40.36} & \textbf{39.27} & \textbf{36.45} & \textbf{36.73} & \textbf{37.69} & \textbf{36.97} \\
& {SSIM} \textuparrow & 0.724 & 0.652 & 0.830 & \textbf{0.903} & \textbf{0.860} & 0.777 & 0.809 & 0.835 & 0.799 \\
& {LPIPS} \textdownarrow & 0.185 & 0.266 & 0.284 & 0.180 & 0.111 & 0.159 & 0.143 & 0.194 & 0.190 \\
\midrule
\multirow{3}{*}{\textbf{Ours}} & {PSNR} \textuparrow & \textbf{34.57} & \textbf{35.15} & 36.02 & 37.59 & 38.04 & 36.19 & 36.09 & 37.36 & 36.38 \\
& {SSIM} \textuparrow & \textbf{0.768} & 0.760 & 0.779 & 0.874 & 0.823 & 0.776 & 0.791 & 0.834 & 0.801 \\
& {LPIPS} \textdownarrow & \textbf{0.159} & \textbf{0.114} &\textbf{0.119} & \textbf{0.051} & \textbf{0.044} & \textbf{0.106} & \textbf{0.097} & \textbf{0.107} & \textbf{0.100} \\
\bottomrule
\end{tabular}
\vspace{-1em}
\label{tab:nvs}
\end{table}
}

\subsection{Novel View Synthesis}

We evaluated our model for novel view synthesis on Replica~\cite{replica19arxiv}. The results can be found in Table~\ref{tab:nvs}. They indicate mixed performance on the different metrics. While NICER-SLAM performs best on SSIM, NeRF-SLAM and our method are almost on par. NeRF-SLAM performs best in PSNR, yet our approach is almost equal in most scenes.
Considering the LPIPS metric, our model shows a significantly superior performance on all scenes, indicating a high subjective similarity for a human observer, underlining the photometric strength of our approach.

\subsection{Runtime Analysis}

We have conducted a runtime and memory utilization analysis of our method. A comparison with concurrent work is presented in Table~\ref{tab:runtime}. The results indicate that our model achieves the highest tracking FPS while leaving the smallest GPU memory footprint. This is primarily due to our runtime- and memory-efficient sparse visual tracking module, and dense geometry enhancement paradigm. In terms of mapping runtime, our method is in the midfield of all competing works. However, it must be emphasized that the two faster methods GO-SLAM~\cite{goslam} and NeRF-SLAM~\cite{rosinol2022nerf} rely on the highly optimized NeRF implementation by Müller et al.\ \cite{mueller2022instant} using custom CUDA kernels, while our work, DIM-SLAM~\cite{li2023dense} and NICER-SLAM~\cite{zhu2023nicerslam} use various unoptimized Python implementations, with our model achieving twice the FPS compared to these Python-based alternatives.
{
\def\arraystretch{1.0}
\setlength\tabcolsep{3.3pt}
\setlength\arrayrulewidth{0.5pt}
\begin{table}[tbp]
\scriptsize
\caption{Runtime and max.\ GPU memory utilization on Replica~\cite{replica19arxiv}. Information on other methods is taken from the respective original papers. "$ \leq $" and "$ \simeq $" indicate missing details on maximum GPU memory usage. In such cases, we have reported the memory capacity of the device used.}
\vspace{-0.5em}
\centering
\begin{tabular}{l|c c c l}
\toprule
\textbf{Model} & \textbf{Tracking} [fps] \textuparrow & \textbf{Mapping} [fps] \textuparrow & \textbf{GPU Memory} [gb] \textdownarrow \\
\midrule
{GO-SLAM} \cite{goslam} & 8 & 8 & 18 \\
{DIM-SLAM} \cite{li2023dense} & 14 & 3 & $ \simeq $11 \\
{NICER-SLAM} \cite{zhu2023nicerslam} & 7 & 2 & $ \leq $40 \\
{NeRF-SLAM} \cite{rosinol2022nerf} & 15 & \textbf{10} & $\simeq$11 \\
 \textbf{Ours} & \textbf{20} & 6 & \textbf{9} \\
\bottomrule
\end{tabular}
\vspace{-1em}
\label{tab:runtime}
\end{table}
}

\subsection{Ablations} \label{ablations}
\PAR{Dense Depths.} We performed an ablation to validate that using aligned dense depths from a monocular estimator does indeed improve performance over directly using the sparse depth patches from our tracking module. The results can be found in Table~\ref{tab:depth_ablation}. The substantial performance improvement from using dense depths indicates that optimizing a NeRF with only a few pixel depths per frame is not sufficient to achieve SOTA results.
{
\def\arraystretch{1.0}
\setlength\tabcolsep{4.2pt}
\setlength\arrayrulewidth{0.5pt}
\begin{table}[tbp]
\scriptsize
\caption{
Ablation study on using dense depth, and surface normal priors for NeRF optimization.
Results on the Replica~\cite{replica19arxiv, Sucar:etal:ICCV2021} dataset.}
\vspace{-0.5em}
\centering
\begin{tabular}{l|c c c c}
\toprule
 & ATE [cm] \textdownarrow & Acc. [cm] \textdownarrow & Com. [cm] \textdownarrow & Rec. [\textless 5cm \%] \textuparrow\\
\midrule
\textbf{Ours} & \textbf{0.43} & \textbf{2.81} & \textbf{3.59} & \textbf{86.3} \\
W/O dense depth & 0.48 & 8.16 & 6.49 & 72.5 \\
W/O normals & 0.45 & 3.22 & 3.79 & 83.7 \\
\bottomrule
\end{tabular}
\vspace{-1em}
\label{tab:depth_ablation}
\end{table}
}
%
% {
% \def\arraystretch{1.0}
% \setlength\tabcolsep{4.9pt}
% \setlength\arrayrulewidth{0.5pt}
% \begin{table}[tbp]
% \scriptsize
% \caption{Impact of using the aligned dense depths from the enhancement module vs.\ directly using the sparse depth from the tracking module. Results on the Replica~\cite{replica19arxiv, Sucar:etal:ICCV2021} dataset.}
% \vspace{-0.5em}
% \centering
% \begin{tabular}{l|c c c c}
% \toprule
%  & ATE [cm] \textdownarrow & Acc. [cm] \textdownarrow & Com. [cm] \textdownarrow & Rec. [\textless 5cm \%] \textuparrow\\
% \midrule
% sparse depth & 0.48 & 8.16 & 6.49 & 72.5 \\
% \textbf{dense depth} & \textbf{0.43} & \textbf{2.81} & \textbf{3.59} & \textbf{86.3} \\
% \bottomrule
% \end{tabular}
% \vspace{-1em}
% \label{tab:depth_ablation}
% \end{table}
% }
\PAR{Surface Normals.} To evaluate the benefit of supervising NeRF optimization with surface normal priors, we performed an ablation by comparing our system to a baseline with the normal consistency loss weight set to zero. The results in Table~\ref{tab:depth_ablation} demonstrate an increase in 3D reconstruction performance by using normal supervision. However, enabling normal supervision slightly increases the runtime, hence one might choose to trade the accuracy gain of normal supervision for a faster runtime.

% {
% \def\arraystretch{1.0}
% \setlength\tabcolsep{5.6pt}
% \setlength\arrayrulewidth{0.5pt}
% \begin{table}[tbp]
% \scriptsize
% \caption{\textcolor{red}{Impact of using surface normal priors for NeRF optimization. Results on the Replica~\cite{replica19arxiv, Sucar:etal:ICCV2021} dataset.}}
% \vspace{-0.5em}
% \centering
% \begin{tabular}{l|c c c c}
% \toprule
%  & ATE [cm] \textdownarrow & Acc. [cm] \textdownarrow & Com. [cm] \textdownarrow & Rec. [\textless 5cm \%] \textuparrow\\
% \midrule
% w/o normals & 0.45 & 3.22 & 3.79 & 83.7 \\
% \textbf{w/ normals} & \textbf{0.43} & \textbf{2.81} & \textbf{3.59} & \textbf{86.3} \\
% \bottomrule
% \end{tabular}
% \vspace{-1em}
% \label{tab:normal_ablation}
% \end{table}
% }

\PAR{Scale Alignment.} As discussed in Sec.~\ref{sec:methodology}, the sparse depths from our tracking module and the dense depths from the monocular estimator are skewed relative to each other, which requires an alignment strategy. In Table~\ref{tab:alignment_ablation} we compare five linear alignment schemes: No alignment of sparse and dense depths (none), alignment based on the minimum and maximum of each depth map (min-max), the least squares alignment proposed by \cite{wofk2023videpth} (least-squares), our probability-based alignment method (ours) as well as its relaxed alternative (see Sec.~\ref{sec:methodology}). Our choice was confirmed as the probability-based alignment performs best.

{
\def\arraystretch{1.0}
\setlength\tabcolsep{3.5pt}
\setlength\arrayrulewidth{0.5pt}
\begin{table}[tbp]
\scriptsize
\caption{Impact of various scale alignment strategies on camera tracking and 3D reconstruction on the Replica~\cite{replica19arxiv, Sucar:etal:ICCV2021} dataset.}
\vspace{-0.5em}
\centering
\begin{tabular}{l|c c c c}
\toprule
 & ATE [cm] \textdownarrow & Acc. [cm] \textdownarrow & Com. [cm] \textdownarrow & Rec. [\textless 5cm \%] \textuparrow \\
\midrule
none & 0.69 & 9.13 & 7.06 & 66.0 \\
min-max & \textbf{0.43} & 8.72 & 9.59 & 65.1 \\
least-squares \cite{wofk2023videpth} & 0.45 & 3.86 & 4.41 & 77.3 \\
ours (relaxed) & 0.45 & 3.57 & 4.04 & 82.9 \\
\textbf{ours} & \textbf{0.43} & \textbf{2.81} & \textbf{3.59} & \textbf{86.3} \\
\bottomrule
\end{tabular}
\vspace{-1em}
\label{tab:alignment_ablation}
\end{table}
}

\iffalse
{
\def\arraystretch{1.0}
\setlength\tabcolsep{2.0pt}
\setlength\arrayrulewidth{0.5pt}
\begin{table}[htbp]
\scriptsize
\caption{Impact of frame skipping on camera tracking and 3D reconstruction on the Replica~\cite{replica19arxiv, Sucar:etal:ICCV2021} dataset.}
\vspace{-0.5em}
\centering
\begin{tabular}{c c|c c c c}
\toprule
Total Frames & Speedup & ATE [cm] \textdownarrow & Acc. [cm] \textdownarrow & Com. [cm] \textdownarrow & Rec. [\textless 5cm \%] \textuparrow\\
\midrule
100\% & $\times$1 & \textbf{0.43} &\textbf{ 2.81} & \textbf{3.59} & 86.3 \\
50\% & $\times$2 & 0.46 & 2.85 & 3.61 & \textbf{86.6} \\
25\% & $\times$4 & 0.50 & 2.83 & 3.70 & 85.9 \\
12.5\% & $\times$8 & 3.88 & 6.34 & 6.13 & 77.1 \\
\bottomrule
\end{tabular}
\vspace{-1em}
\label{tab:skip_ablation}
\end{table}
}

\PAR{Frame Skipping.} Analogous to the authors of GO-SLAM~\cite{goslam}, we performed a frame skipping ablation to assess what percentage of the total frames is necessary to still achieve SOTA performance. This has a direct impact on the runtime and performance of our model.
The results are shown in Table~\ref{tab:skip_ablation}. We discovered that, even with 25\% of all frames, our approach still achieves nearly equivalent results, with performance degrading only after a reduction to 12.5\% (every 8th frame). This means that skipping 75\% of the frames could lead to a fourfold speedup of the model, allowing its use in real-time applications.
\fi
	\section{Conclusion}
\label{sec:conclusion}
We introduce NeRF-VO, a novel neural visual odometry system that combines a learning-based sparse visual odometry for pose tracking, a monocular prediction network for inferring dense geometry cues, and a neural radiance field optimization for pose and dense geometry refinement. NeRF-VO surpasses SOTA methods, demonstrating superior pose estimation accuracy and delivering high-quality dense mapping, all while maintaining low pose tracking latency and GPU memory consumption. As part of future research, it would be compelling to integrate visual or geometric constraints obtained from the neural scene representation into the pose tracking front-end. This integration has the potential to further reduce drift and improve the accuracy of initial pose tracking.
	
	\bibliographystyle{IEEEtran}
	\bibliography{IEEEabrv,main}

\begin{thebibliography}{10}
\providecommand{\url}[1]{#1}
\csname url@samestyle\endcsname
\providecommand{\newblock}{\relax}
\providecommand{\bibinfo}[2]{#2}
\providecommand{\BIBentrySTDinterwordspacing}{\spaceskip=0pt\relax}
\providecommand{\BIBentryALTinterwordstretchfactor}{4}
\providecommand{\BIBentryALTinterwordspacing}{\spaceskip=\fontdimen2\font plus
\BIBentryALTinterwordstretchfactor\fontdimen3\font minus
  \fontdimen4\font\relax}
\providecommand{\BIBforeignlanguage}[2]{{%
\expandafter\ifx\csname l@#1\endcsname\relax
\typeout{** WARNING: IEEEtran.bst: No hyphenation pattern has been}%
\typeout{** loaded for the language `#1'. Using the pattern for}%
\typeout{** the default language instead.}%
\else
\language=\csname l@#1\endcsname
\fi
#2}}
\providecommand{\BIBdecl}{\relax}
\BIBdecl

\bibitem{replica19arxiv}
J.~Straub, T.~Whelan, L.~Ma, Y.~Chen, E.~Wijmans, S.~Green, J.~J. Engel,
  R.~Mur-Artal, C.~Ren, S.~Verma, A.~Clarkson, M.~Yan, B.~Budge, Y.~Yan,
  X.~Pan, J.~Yon, Y.~Zou, K.~Leon, N.~Carter, J.~Briales, T.~Gillingham,
  E.~Mueggler, L.~Pesqueira, M.~Savva, D.~Batra, H.~M. Strasdat, R.~D. Nardi,
  M.~Goesele, S.~Lovegrove, and R.~Newcombe, ``The {Replica} dataset: A digital
  replica of indoor spaces,'' 2019.

\bibitem{Sucar:etal:ICCV2021}
E.~Sucar, S.~Liu, J.~Ortiz, and A.~J. Davison, ``{iMAP: I}mplicit mapping and
  positioning in real-time,'' in \emph{ICCV}, 2021.

\bibitem{mildenhall2020nerf}
B.~Mildenhall, P.~P. Srinivasan, M.~Tancik, J.~T. Barron, R.~Ramamoorthi, and
  R.~Ng, ``Nerf: Representing scenes as neural radiance fields for view
  synthesis,'' in \emph{ECCV}, 2020.

\bibitem{Chen2022ECCV}
A.~Chen, Z.~Xu, A.~Geiger, J.~Yu, and H.~Su, ``{TensoRF}: Tensorial radiance
  fields,'' in \emph{ECCV}, 2022.

\bibitem{yu2022plenoxels}
S.~Fridovich-Keil, A.~Yu, M.~Tancik, Q.~Chen, B.~Recht, and A.~Kanazawa,
  ``Plenoxels: Radiance fields without neural networks,'' in \emph{CVPR}, 2022.

\bibitem{mueller2022instant}
T.~M\"{u}ller, A.~Evans, C.~Schied, and A.~Keller, ``Instant neural graphics
  primitives with a multiresolution hash encoding,'' \emph{ACM Trans. Graph.},
  2022.

\bibitem{engel2014lsd}
J.~Engel, T.~Sch{\"o}ps, and D.~Cremers, ``Lsd-slam: Large-scale direct
  monocular slam,'' in \emph{ECCV}, 2014.

\bibitem{engel2017direct}
J.~Engel, V.~Koltun, and D.~Cremers, ``Direct sparse odometry,'' \emph{TPAMI},
  2018.

\bibitem{zubizarreta2020direct}
J.~Zubizarreta, I.~Aguinaga, and J.~M.~M. Montiel, ``Direct sparse mapping,''
  \emph{IEEE Transactions on Robotics}, 2020.

\bibitem{davison2007monoslam}
A.~J. Davison, I.~D. Reid, N.~D. Molton, and O.~Stasse, ``Monoslam: Real-time
  single camera slam,'' \emph{TPAMI}, 2007.

\bibitem{campos2021orb}
C.~Campos, R.~Elvira, J.~J.~G. Rodríguez, J.~M. M.~Montiel, and J.~D.~Tardós,
  ``Orb-slam3: An accurate open-source library for visual, visual–inertial,
  and multimap slam,'' \emph{IEEE Transactions on Robotics}, 2021.

\bibitem{cvivsic2022soft2}
I.~Cvišić, I.~Marković, and I.~Petrović, ``Soft2: Stereo visual odometry
  for road vehicles based on a point-to-epipolar-line metric,'' \emph{IEEE
  Transactions on Robotics}, 2023.

\bibitem{niceslam}
Z.~Zhu, S.~Peng, V.~Larsson, W.~Xu, H.~Bao, Z.~Cui, M.~R. Oswald, and
  M.~Pollefeys, ``{NICE-SLAM}: Neural implicit scalable encoding for slam,'' in
  \emph{CVPR}, 2022.

\bibitem{pointslam}
E.~Sandstr\"om, Y.~Li, L.~Van~Gool, and M.~R. Oswald, ``Point-slam: Dense
  neural point cloud-based slam,'' in \emph{ICCV}, 2023.

\bibitem{coslam}
H.~Wang, J.~Wang, and L.~Agapito, ``{Co-SLAM}: Joint coordinate and sparse
  parametric encodings for neural real-time slam,'' in \emph{CVPR}, 2023.

\bibitem{voxfusion}
X.~Yang, H.~Li, H.~Zhai, Y.~Ming, Y.~Liu, and G.~Zhang, ``{Vox-Fusion}: Dense
  tracking and mapping with voxel-based neural implicit representation,'' in
  \emph{ISMAR}, 2022.

\bibitem{featurerealisticneural}
K.~Mazur, E.~Sucar, and A.~J. Davison, ``Feature-realistic neural fusion for
  real-time, open set scene understanding,'' in \emph{ICRA}, 2023.

\bibitem{rosinol2022nerf}
A.~Rosinol, J.~J. Leonard, and L.~Carlone, ``Nerf-slam: Real-time dense
  monocular slam with neural radiance fields,'' in \emph{IROS}, 2023.

\bibitem{zhu2023nicerslam}
Z.~Zhu, S.~Peng, V.~Larsson, Z.~Cui, M.~R. Oswald, A.~Geiger, and M.~Pollefeys,
  ``Nicer-slam: Neural implicit scene encoding for rgb slam,'' in \emph{3DV},
  2024.

\bibitem{li2023dense}
H.~Li, X.~Gu, W.~Yuan, L.~Yang, Z.~Dong, and P.~Tan, ``Dense rgb slam with
  neural implicit maps,'' in \emph{ICLR}, 2023.

\bibitem{orbeezslam}
C.-M. Chung, Y.-C. Tseng, Y.-C. Hsu, X.-Q. Shi, Y.-H. Hua, J.-F. Yeh, W.-C.
  Chen, Y.-T. Chen, and W.~H. Hsu, ``Orbeez-slam: A real-time monocular visual
  slam with orb features and nerf-realized mapping,'' in \emph{ICRA}, 2023.

\bibitem{goslam}
Y.~Zhang, F.~Tosi, S.~Mattoccia, and M.~Poggi, ``{GO-SLAM}: Global optimization
  for consistent 3d instant reconstruction,'' in \emph{ICCV}, 2023.

\bibitem{hislam}
W.~Zhang, T.~Sun, S.~Wang, Q.~Cheng, and N.~Haala, ``{HI-SLAM}: Monocular
  real-time dense mapping with hybrid implicit fields,'' in \emph{IEEE Robotics
  and Automation Letters}, 2024.

\bibitem{survey}
F.~Tosi, Y.~Zhang, Z.~Gong, E.~Sandström, S.~Mattoccia, M.~R. Oswald, and
  M.~Poggi, ``How nerfs and 3d gaussian splatting are reshaping slam: a
  survey,'' 2024.

\bibitem{cnnslam}
K.~Tateno, F.~Tombari, I.~Laina, and N.~Navab, ``{CNN-SLAM}: Real-time dense
  monocular slam with learned depth prediction,'' in \emph{CVPR}, 2017.

\bibitem{yang2020d3vo}
N.~Yang, L.~v. Stumberg, R.~Wang, and D.~Cremers, ``{D3VO}: Deep depth, deep
  pose and deep uncertainty for monocular visual odometry,'' in \emph{CVPR},
  2020.

\bibitem{wang2021tartanvo}
W.~Wang, Y.~Hu, and S.~Scherer, ``{TartanVO}: A generalizable learning-based
  vo,'' in \emph{CoRL}, 2021.

\bibitem{teed2021droid}
Z.~Teed and J.~Deng, ``{DROID-SLAM}: Deep visual slam for monocular, stereo,
  and rgb-d cameras,'' in \emph{Advances in Neural Information Processing
  Systems}, 2021.

\bibitem{teed2022deep}
Z.~Teed, L.~Lipson, and J.~Deng, ``Deep patch visual odometry,'' \emph{Advances
  in Neural Information Processing Systems}, 2023.

\bibitem{dtam}
R.~A. Newcombe, S.~J. Lovegrove, and A.~J. Davison, ``{DTAM}: Dense tracking
  and mapping in real-time,'' in \emph{ICCV}, 2011.

\bibitem{deepfactors}
J.~Czarnowski, T.~Laidlow, R.~Clark, and A.~J. Davison, ``{DeepFactors}:
  Real-time probabilistic dense monocular slam,'' \emph{IEEE Robotics and
  Automation Letters}, 2020.

\bibitem{deeptam}
H.~Zhou, B.~Ummenhofer, and T.~Brox, ``{DeepTAM}: Deep tracking and mapping,''
  in \emph{ECCV}, 2018.

\bibitem{demon}
B.~Ummenhofer, H.~Zhou, J.~Uhrig, N.~Mayer, E.~Ilg, A.~Dosovitskiy, and
  T.~Brox, ``{DeMoN}: Depth and motion network for learning monocular stereo,''
  in \emph{CVPR}, 2017.

\bibitem{codeslam}
M.~Bloesch, J.~Czarnowski, R.~Clark, S.~Leutenegger, and A.~J. Davison,
  ``{CodeSLAM} — learning a compact, optimisable representation for dense
  visual slam,'' in \emph{CVPR}, 2018.

\bibitem{zuo2021codevio}
X.~Zuo, N.~Merrill, W.~Li, Y.~Liu, M.~Pollefeys, and G.~Huang, ``{CodeVIO:
  V}isual-inertial odometry with learned optimizable dense depth,'' in
  \emph{ICRA}, 2021.

\bibitem{tandem}
L.~Koestler, N.~Yang, N.~Zeller, and D.~Cremers, ``{TANDEM:} tracking and dense
  mapping in real-time using deep multi-view stereo,'' in \emph{CoRL}, 2021.

\bibitem{xin2023simplemapping}
Y.~Xin, X.~Zuo, D.~Lu, and S.~Leutenegger, ``{SimpleMapping: Real-Time
  Visual-Inertial Dense Mapping with Deep Multi-View Stereo},'' \emph{ISMAR},
  2023.

\bibitem{NEWTON}
H.~Matsuki, K.~Tateno, M.~Niemeyer, and F.~Tombari, ``{NEWTON}: Neural
  view-centric mapping for on-the-fly large-scale slam,'' \emph{IEEE Robotics
  and Automation Letters}, 2024.

\bibitem{Lisus2023TowardsOW}
D.~Lisus, C.~Holmes, and S.~Waslander, ``Towards open world nerf-based slam,''
  in \emph{CRV}, 2023.

\bibitem{3DGS}
B.~Kerbl, G.~Kopanas, T.~Leimkuehler, and G.~Drettakis, ``3d gaussian splatting
  for real-time radiance field rendering,'' \emph{ACM Trans. Graph.}, 2023.

\bibitem{GSSLAM}
H.~Matsuki, R.~Murai, P.~H.~J. Kelly, and A.~J. Davison, ``{G}aussian
  {S}platting {SLAM},'' in \emph{CVPR}, 2024.

\bibitem{GS-SLAM}
C.~Yan, D.~Qu, D.~Wang, D.~Xu, Z.~Wang, B.~Zhao, and X.~Li, ``Gs-slam: Dense
  visual slam with 3d gaussian splatting,'' in \emph{CVPR}, 2024.

\bibitem{SplaTAM}
N.~Keetha, J.~Karhade, K.~M. Jatavallabhula, G.~Yang, S.~Scherer, D.~Ramanan,
  and J.~Luiten, ``Splatam: Splat, track \& map 3d gaussians for dense rgb-d
  slam,'' in \emph{CVPR}, 2024.

\bibitem{Photo-SLAM}
H.~Huang, L.~Li, C.~Hui, and S.-K. Yeung, ``Photo-slam: Real-time simultaneous
  localization and photorealistic mapping for monocular, stereo, and rgb-d
  cameras,'' in \emph{CVPR}, 2024.

\bibitem{nerfstudio}
M.~Tancik, E.~Weber, E.~Ng, R.~Li, B.~Yi, T.~Wang, A.~Kristoffersen, J.~Austin,
  K.~Salahi, A.~Ahuja, D.~Mcallister, J.~Kerr, and A.~Kanazawa, ``Nerfstudio: A
  modular framework for neural radiance field development,'' in
  \emph{SIGGRAPH}, 2023.

\bibitem{eftekhar2021omnidata}
A.~Eftekhar, A.~Sax, J.~Malik, and A.~Zamir, ``Omnidata: A scalable pipeline
  for making multi-task mid-level vision datasets from 3d scans,'' in
  \emph{ICCV}, 2021.

\bibitem{Ranftl2020}
R.~Ranftl, K.~Lasinger, D.~Hafner, K.~Schindler, and V.~Koltun, ``Towards
  robust monocular depth estimation: Mixing datasets for zero-shot
  cross-dataset transfer,'' \emph{TPAMI}, 2022.

\bibitem{Ranftl2021}
R.~Ranftl, A.~Bochkovskiy, and V.~Koltun, ``Vision transformers for dense
  prediction,'' in \emph{ICCV}, 2021.

\bibitem{depthloss}
K.~Deng, A.~Liu, J.-Y. Zhu, and D.~Ramanan, ``Depth-supervised nerf: Fewer
  views and faster training for free,'' in \emph{CVPR}, 2022.

\bibitem{normalloss}
Z.~Yu, S.~Peng, M.~Niemeyer, T.~Sattler, and A.~Geiger, ``{MonoSDF}: Exploring
  monocular geometric cues for neural implicit surface reconstruction,'' in
  \emph{Advances in Neural Information Processing Systems}, 2022.

\bibitem{interlevelloss}
J.~T. Barron, B.~Mildenhall, D.~Verbin, P.~P. Srinivasan, and P.~Hedman,
  ``{Mip-NeRF 360}: Unbounded anti-aliased neural radiance fields,'' in
  \emph{CVPR}, 2022.

\bibitem{dai2017scannet}
A.~Dai, A.~X. Chang, M.~Savva, M.~Halber, T.~Funkhouser, and M.~Niessner,
  ``{ScanNet}: Richly-annotated 3d reconstructions of indoor scenes,'' in
  \emph{CVPR}, 2017.

\bibitem{dai2017bundlefusion}
A.~Dai, M.~Nie\ss{}ner, M.~Zollh\"{o}fer, S.~Izadi, and C.~Theobalt,
  ``{BundleFusion}: Real-time globally consistent 3d reconstruction using
  on-the-fly surface reintegration,'' \emph{ACM Trans. Graph.}, 2017.

\bibitem{shotton2013scene}
J.~Shotton, B.~Glocker, C.~Zach, S.~Izadi, A.~Criminisi, and A.~Fitzgibbon,
  ``Scene coordinate regression forests for camera relocalization in rgb-d
  images,'' in \emph{CVPR}, 2013.

\bibitem{kinectfusion}
R.~A. Newcombe, S.~Izadi, O.~Hilliges, D.~Molyneaux, D.~Kim, A.~J. Davison,
  P.~Kohi, J.~Shotton, S.~Hodges, and A.~Fitzgibbon, ``{KinectFusion}:
  Real-time dense surface mapping and tracking,'' in \emph{ISMAR}, 2011.

\bibitem{leutenegger2022okvis2}
S.~Leutenegger, ``{OKVIS2: R}ealtime scalable visual-inertial slam with loop
  closure,'' 2022.

\bibitem{Kabsch:a15629}
W.~Kabsch, ``{A discussion of the solution for the best rotation to relate two
  sets of vectors},'' \emph{Acta Crystallographica Section A}, 1978.

\bibitem{88573}
S.~Umeyama, ``Least-squares estimation of transformation parameters between two
  point patterns,'' \emph{TPAMI}, 1991.

\bibitem{sturm12iros}
J.~Sturm, N.~Engelhard, F.~Endres, W.~Burgard, and D.~Cremers, ``A benchmark
  for the evaluation of rgb-d slam systems,'' in \emph{IROS}, 2012.

\bibitem{ssim}
Z.~Wang, A.~Bovik, H.~Sheikh, and E.~Simoncelli, ``Image quality assessment:
  from error visibility to structural similarity,'' \emph{IEEE Transactions on
  Image Processing}, 2004.

\bibitem{lpips}
R.~Zhang, P.~Isola, A.~A. Efros, E.~Shechtman, and O.~Wang, ``The unreasonable
  effectiveness of deep features as a perceptual metric,'' in \emph{CVPR},
  2018.

\bibitem{wofk2023videpth}
D.~Wofk, R.~Ranftl, M.~Müller, and V.~Koltun, ``Monocular visual-inertial
  depth estimation,'' in \emph{ICRA}, 2023.

\end{thebibliography}
	
	% \clearpage
	% \input{sections/6_rebuttal}
	
\end{document}